\documentclass[10pt, conference, letterpaper]{IEEEtran}

\usepackage{microtype}
\usepackage{graphicx}
\usepackage{subfigure}
\usepackage{booktabs} 

\usepackage{algorithm}
\usepackage{algpseudocode}
\usepackage{caption}
\usepackage{hyperref}

\usepackage{amsmath}
\usepackage{amssymb}
\usepackage{mathtools}
\usepackage{amsthm}

\usepackage[capitalize,noabbrev]{cleveref}

\theoremstyle{plain}

\theoremstyle{definition}

\theoremstyle{remark}

\usepackage[textsize=tiny]{todonotes}

\begin{document}

\title{Model-Distributed Inference for Large Language Models at the Edge}

\author{\IEEEauthorblockN{Davide Macario}
\IEEEauthorblockA{
\textit{University of Illinois Chicago}\\
dmacar4@uic.edu}
\and
\IEEEauthorblockN{Hulya Seferoglu}
\IEEEauthorblockA{
\textit{University of Illinois Chicago}\\
hulya@uic.edu}
\and
\IEEEauthorblockN{Erdem Koyuncu}
\IEEEauthorblockA{
\textit{University of Illinois Chicago}\\
ekoyuncu@uic.edu}
}

\maketitle

\begin{abstract}

We introduce Model-Distributed Inference for Large-Language Models (MDI-LLM), a novel framework designed to facilitate the deployment of state-of-the-art large-language models (LLMs) across low-power devices at the edge. This is accomplished by dividing the model into multiple partitions, which are then assigned to different devices/nodes within the network. These nodes exchange intermediate activation vectors via device-to-device links, enabling collaborative computation. To enhance the efficiency of this process, we propose the ``recurrent pipeline parallelism'' technique, which reduces idle time on each device and facilitates parallel inference during the generation of multiple text sequences. By leveraging the combined computational resources of multiple edge devices, MDI-LLM enables the deployment of LLMs that exceed the memory capacity of individual devices, making it possible to perform inference on low-cost hardware. Furthermore, as the number of participating devices increases, MDI-LLM boosts token generation throughput and reduces memory consumption per device.

\end{abstract}

\section{Introduction}\label{intro}
Over the past decade, generative models have risen to become a mainstream application of Artificial Intelligence (AI) and Machine Learning (ML). The main direction that is being pursued by research and industry, however, is that of
\textit{scaling-up}, i.e., by increasing both the size of the models and
the capabilities of the hardware to achieve better performance.
In the case of Large Language Models (LLMs) for text generation, perhaps the most ubiquitous form of generative AI, the size of the deep neural networks grew from
the order of hundreds of millions of parameters (GPT-2 \cite{radford2019language}) to that of hundreds of billions
of parameters (GPT-3 175 B \cite{brown2020language}, PaLM 540 B \cite{chowdhery2022palm}).
This translates into the need for a huge amount of computation power, making
it impossible, or very expensive, to deploy LLMs outside the cloud.
In particular, generative AI is still unexplored for applications at the edge, as the corresponding devices typically lack the capabilities to run LLMs independently and have to rely on a remote cloud. 

Deploying AI/ML models at the edge can be desirable for many reasons, including privacy, availability, and the rising cost of cloud resources \cite{10.1145/3555802}.
Thus, many approaches have been devised in the literature for performing traditional deep neural
network inference at the edge \cite{chen2019, 10.1145/3555802}. The most common ones involve forms of parallelism to split the computational load among multiple interconnected edge devices.
When discussing parallelism in deep learning, the three main approaches are tensor parallelism \cite{shoeybi2020megatronlm, Wang_2022}, data parallelism \cite{chen2019efficient}, and model/pipeline parallelism \cite{DBLP:journals/corr/abs-1811-06965, hu2021pipeline, li2021respipe}.
Out of these three, only data parallelism and model parallelism have been successfully applied at the edge. 
Tensor parallelism requires intensive communication and syncing between the nodes participating in it \cite{shoeybi2020megatronlm}, making it unsuitable for physically distributed applications.

In data parallelism, the idea is to split the data that needs to be processed among a set of nodes so that they can work in parallel on distinct samples to reduce the overall processing time.
This approach is particularly suited for applications involving processing large amounts of data, as adding nodes to the network increases the system's processing capability at the expense of the communication cost required to transmit the input samples to each node.
The main drawback of data parallelism is the need to fit the full model on every network device, which limits the range of applications that can benefit from this approach.

Model parallelism, instead, enables machine learning model deployment by partitioning the model and offloading the model parts over multiple computing nodes.
This technique was initially designed for model training in the cloud \cite{DBLP:journals/corr/abs-1811-06965}, but then used for machine learning inference at the edge \cite{10138654,li2023model,colocrese2024early,li2024priority}.
Processing data using this approach requires the compute nodes to exchange the model's intermediate activations while only processing a subset of the model layers, forming a communication chain between the network nodes.
To achieve parallel computation, model parallelism is paired with ``pipeline parallelism'' \cite{hu2021pipeline}, which consists of processing the samples consecutively one after the other, preventing idle nodes, by beginning processing a new sample as soon as the previous one is transmitted to the next node.
Model parallelism reduces the memory requirements as compared to data parallelism thanks to partitioning the model.

In this paper, we consider using an existing infrastructure of low-cost edge devices collaboratively and distributedly for LLM inference. As compared to existing model-distributed inference (MDI) works \cite{10138654,li2023model,colocrese2024early,li2024priority}, model-distributed LLMs are challenging due to their autoregressive nature, where models in which text is generated by repeatedly feeding back the output into the input. We achieve this challenge in this paper and design Model-Distributed Inference for Large-Language Models (MDI-LLM) by designing a technique that we name ``recurrent pipeline parallelism.''
%
Another key consideration when working with LLMs is ensuring strong overall performance, particularly in terms of generation speed—measured by the number of tokens produced per unit of time. To achieve this, our MDI-LLM design incorporates techniques such as KV caching \cite{pope2022efficiently} and Grouped Query Attention (GQA) \cite{ainslie2023gqa}. While KV caching and GQA were originally developed to accelerate LLM inference in centralized settings, we adapt these techniques to a distributed environment in our MDI-LLM design

The rest of this paper is organized as follows. Section \ref{sec:related} provides related work. Section \ref{sec:MDI-LLM} presents our MDI-LLM framework. Section \ref{sec:eval} provides the performance analysis of MDI-LLM. Section \ref{sec:conc} concludes the paper. 
%


\section{Related Work\label{sec:related}}


Recent research trends have focused on making large language models (LLMs) more accessible and less resource-intensive. The most common approach is that of model quantization \cite{gholami2021survey, zhao2024atom}.  
This approach enables a reduction in the memory footprint of large models by quantizing their parameters to lower-bit representations.
%
Reducing the precision of model representations can accelerate inference while typically preserving model performance \cite{jaiswal2024compressing}.
However, it requires hardware support to execute operations with the same precision, and many accelerators, especially older ones or those not specifically created for machine learning tasks, are not optimized to work with different representations.
It can also be noted that since weight quantization does not reduce the number of model parameters, some LLMs may still be too large to run on a single computer. Other approaches, such as model pruning \cite{sun2024simple} and knowledge distillation \cite{gou2021knowledge}, are also being investigated as possible ways to reduce the model size and, consequently, the hardware and energy requirements to run LLMs.

To fully leverage the capabilities of resource-constrained environments, model-distributed inference (MDI) with multiple split points has been proposed. The model partitioning problem for MDI is addressed in \cite{EdgePipe-hu2021pipeline} using a dynamic programming approach. To accommodate the heterogeneous and time-varying nature of edge resources, an adaptive and resilient layer allocation strategy is introduced in \cite{10138654}. Model-distributed inference across multiple input sources is explored in \cite{li2023model}, while \cite{colocrese2024early} incorporates early-exit mechanisms into the MDI framework to optimize latency and efficiency. Model-distributed inference for multi-modal inputs, where each modality may have distinct priorities, is considered in \cite{li2024priority}. In contrast to these prior works, we consider MDI for LLMs. A distributed LLM framework that splits model layers across edge devices with limited compute resources is considered in \cite{perello2024jarvis}. As compared to this work, our MDI-LLM uses recurrent pipeline parallelism and incorporates techniques such as KV caching and Grouped Query Attention (GQA), which are crucial for the efficiency of MDI-LLM. 


\section{\label{sec:MDI-LLM}The MDI-LLM Framework}


\subsection{System Architecture}

We consider a set of ``virtual'' nodes that can be deployed on separate devices.
Two types of nodes are defined: ``starter'' nodes and ``secondary'' nodes, with slightly different functions and model chunk structures.
At runtime, these nodes will form a ``ring'' overlay network by creating direct low-level connections over TCP/IP to exchange the intermediate activations sequentially and allow the output tokens to be fed back into the input to continue generation.
All nodes also act as HTTP servers to exchange coordination information at the beginning and end of the generation.

The ``starter'' node is the application's entry point.
It accepts the input prompt and returns the generated text.
When deploying a model, it is always necessary to define one starter node, which will act as a coordinator for the whole network by initializing all secondary nodes present and assigning them their own chunk and the information about their neighbors in the overlay for message transmission.
Notice that starter nodes are also responsible for processing a local model chunk.
They don't just have a role as coordinators but are active entities in the inference process.

``Secondary'' nodes, instead, are the ``worker'' nodes of the system.
They are completely agnostic of the overall inference process, as their only task is to receive an input vector for their own local model chunk from the previous node in the message chain, process it, and transmit the output to the next model in the node.
They receive set-up information over HTTP from the starter node and stop working only when the same node advertises the end of the generation process.
Adding more secondary nodes to the system allows larger LLMs and more samples to be processed.

A crucial aspect of this implementation is the exchange of messages containing the intermediate activations.
As anticipated, connections are created with the previous and next nodes in the message chain when the system is set up.
For each inference run, the nodes will keep the same connections open to prevent having to recreate them every time a message is exchanged.
These channels are directly built over TCP/IP (using Python sockets) to minimize the overhead while still achieving reliable data exchange thanks to TCP.
At the application layer, the only protocol used consists of adding a header to each message to provide information about the payload size and prevent issues related to truncated messages.


Our design uses different threads for processing the local model, receiving inputs, and transmitting outputs.
To allow communication between the threads, we implemented two FIFO message queues, one for incoming and one for outgoing.
The messages that arrive at the node are received on one thread, which continuously listens on the socket connecting the node to the previous neighbor.
These messages are then placed in the input message queue, where the main ``processing'' thread extracts them one at a time, forwards the contents through the local model layers, and creates the message containing the output activations, which is then placed in the output queue.
The third thread, which handles communication with the next node in the chain, will empty the output queue.



\subsection{Model Partitioning}



An essential component of model-distributed inference is the partitioning of the model across multiple nodes. This process involves several considerations when determining how to allocate layers to individual devices. A primary objective is to prevent bottlenecks caused by slower nodes by ensuring that the processing time is balanced across the system. This typically requires assigning layers in proportion to each node’s computational capabilities. Numerous strategies have been proposed in the literature to optimize this layer-to-node assignment and achieve balanced workload distribution \cite{10.1145/3656041, 10138654}

\begin{figure}[t]
    \centering
    \scalebox{0.25}{
    \includegraphics[angle=90]{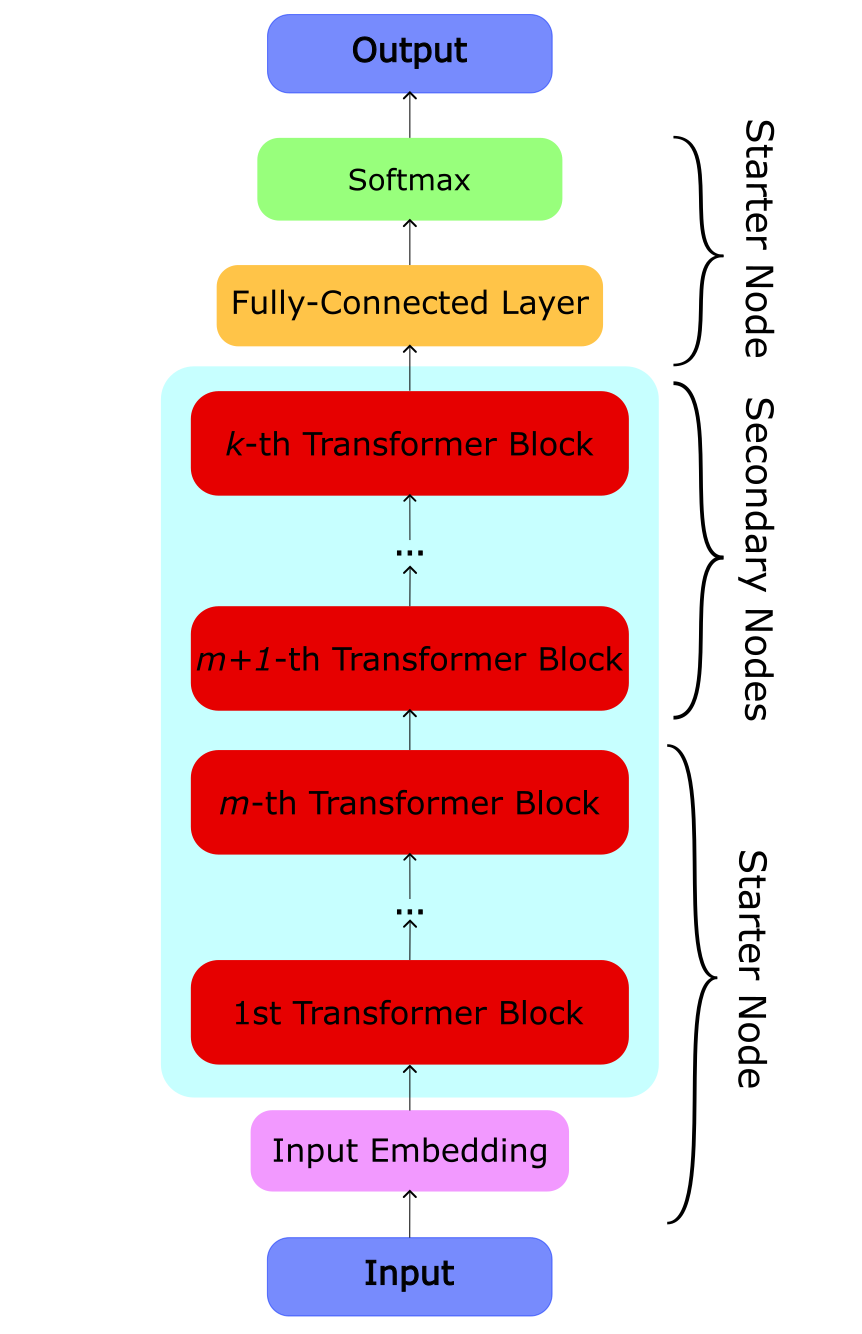}}
    \vspace{-35pt}
    \caption{Model partitioning scheme of MDI-LLM.}
    \label{model-partition}
    \vspace{-10pt}
\end{figure}

Our main contribution in the context of model partitioning focuses on identifying the most effective method for dividing the layers of a transformer-based, decoder-only neural network, with the goal of eliminating communication bottlenecks that can arise during distributed inference.
LLMs are composed of a series of blocks (the ``transformer blocks'') that are identical in size and have the same input and output dimensions (the ``embedding dimension''). A transformer block is itself a cascade of a Multi-Head Attention (MHA), and a fully-connected layer, together with a residual connection. The only layer that works with a different dimension is the output fully connected layer, which provides the output logits, i.e., a vector having a length equal to the vocabulary size (i.e., the number of possible tokens). This vector is used to obtain a probability mass function that is then sampled to extract the next token in the sequence. The model partition scheme of MDI-LLM is shown in  Fig. \ref{model-partition}, where it assigns to the starter node both the initial and final layers of the transformer decoder alongside a few of the first transformer layers, while the secondary nodes all contain a number of transformer blocks (that in general should reflect the computation capabilities of each specific device).


\subsection{Recurrent Pipeline Parallelism}

One of the primary contributions of this work is to implement pipeline parallelism for recurrent neural network models such as LLMs. 
Our ``recurrent pipeline parallelism'' technique allows the network of nodes to process multiple samples, i.e., pieces of text, in parallel, resulting in a reduced generation time compared to sequential generation.
Additionally, if the number of samples is greater than or equal to the number of nodes and the model is partitioned in a balanced way (i.e., local processing takes the same amount of time on every node), it is possible to prevent having idle nodes during generation.

\begin{figure}[t]
    \centering
    \scalebox{0.25}{
    \includegraphics{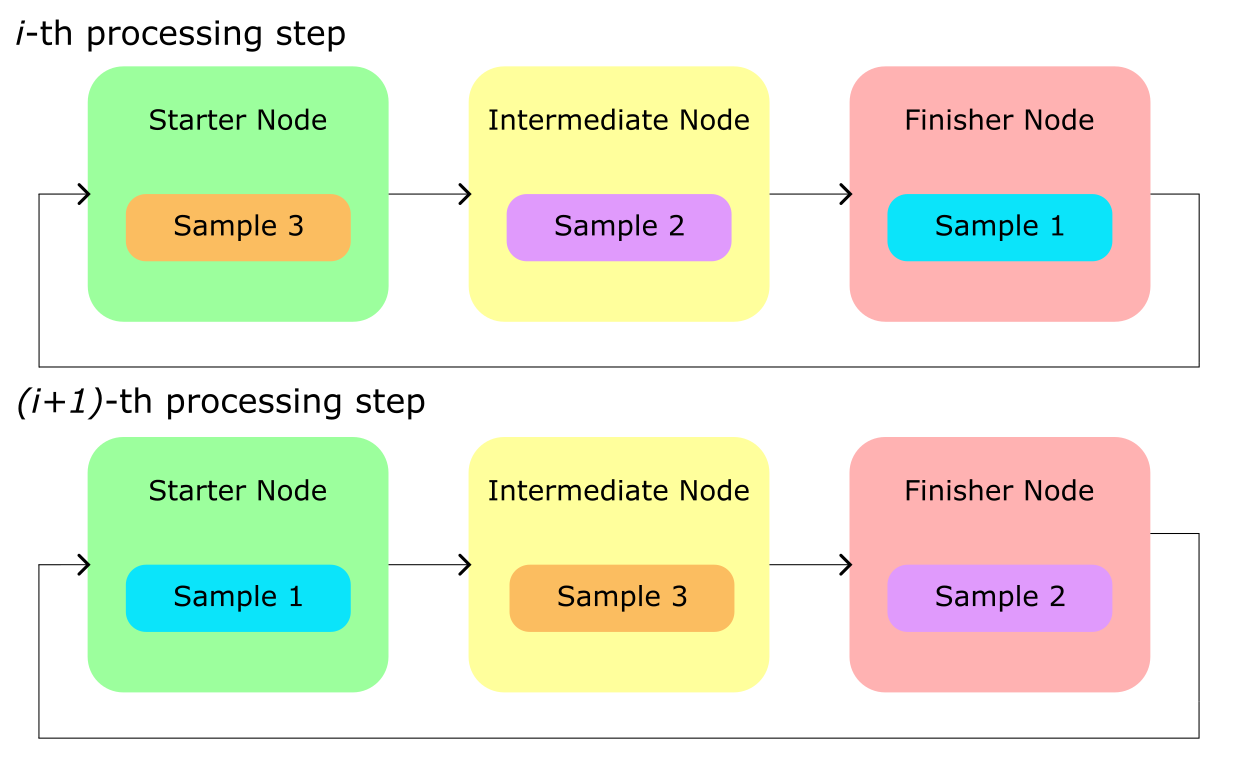}}
    \vspace{-5pt}
    \caption{Recurrent pipeline parallelism for MDI-LLM.}
    \label{pipeline-parallelism}
    \vspace{-10pt}
\end{figure}

Fig. \ref{pipeline-parallelism} illustrates how recurrent pipeline parallelism works in practice.
Considering a network of three nodes generating three pieces of text, the first node (the starter) will start processing sample 1 using its local model chunk and send the output to the following node (the first ``secondary'' node in the communication chain).
As soon as it sends the output to the following node, the starter begins processing sample number 2.
Assuming the processing time is the same on different devices, both the starter and the first secondary node will obtain the output of their model chunk simultaneously. 
Then, they will forward the result to the next node, and the starter node will start processing the third sample, while the secondary nodes will be processing the second and the first samples, respectively. 
At the next iteration, the last node will transmit its output (sample 1) back to the starter node, passing it through the final model layers to obtain the first ``new'' token for the first sample.
After obtaining the new token, the starter node will use it as input for another forward pass, repeating the same cycle. 


\subsection{Rotating KV Caches}
Since large LLMs support a longer context and use a larger embedding dimension, simply scaling a smaller system up results in a huge message size during inference. Without KV caching, each transmitted message includes all token embeddings in the context, i.e., a tensor of size $(\text{context length})\times (\text{embedding dimension})$.
The slowdown caused by the message transmission also adds to the slower processing at each node since the K and V matrices at every self-attention layer have to be computed from scratch at each forward pass.

To solve this issue, we have included KV caching to our model to incorporate large LLMs.
This enabled smaller messages (as only the last token embedding needs to be propagated through the neural network) and faster processing (as large matrix multiplications are not needed to evaluate the Key and Value matrices but only to extend the cached matrices with the information from the new token).
Due to the use of recurrent pipeline parallelism, for which at each node we process different pieces of text one after the other, we opted for storing the cached Key and Value matrices for each of the samples separately and swap the ``active'' ones every time a different sample is processed.

\subsection{MDI-LLM Operation}

In this section, we provide the algorithms the nodes use at runtime.
Algorithm \ref{alg:starter} shows the main loop for the starter node.
It assumes the node has already set up the other nodes via the HTTP coordination channel and created the message queues.
A key aspect is that the loop operation changes based on the iteration number.
If the current iteration is the first iteration for the sample extracted from the input queue, the starter node will initialize the KV cache for the node and forward the full sample (all encoded tokens) through the initial model layers to build up the cache.

Notice that, due to the partition scheme, the output model portion, i.e., the one that yields the output probability mass function over the tokens, does not contain transformer blocks and, therefore, does not require KV cache initialization, making it easier to handle.
Once we are in the final iteration for the current sample, i.e., we are producing the last token, instead, the starter node will only apply the output layers to the sample without feeding it back to the input and, most importantly, without placing the result in the output message queue.

\begin{algorithm}[tb]
   \caption{Starter node main processing loop}
   \label{alg:starter}
\begin{algorithmic}
    \State \textbf{input:} $n\_samples$, $n\_tokens$
    \State Initialize sockets and message queues;
    \State Load model chunk and initialize model state;
    \State Load tokenizer;
    \State Encode prompts (one for each sample);
    \State Place input tensors in input message queue;
    \State $n\_iterations$ = $n\_samples \times n\_tokens$;
    \For{$k=1$ {\bfseries to} $n\_iterations$}
        \State Extract sample $s_n$ ($n = k \mod n\_samples$) from input queue;
        
        \If{$k < n\_samples$}
            \State \# \textit{First iteration for each sample}
            \State \verb|init_KV_cache(|$sample=s_n$\verb|)|;
        \Else
            \State Process output layers;
            \State Sample output probability distribution and get token;
            \State Append new token to the current sample $s_n$;
        \EndIf

        \If{$k < n\_samples \times (n\_tokens - 1)$}
            \State Activate KV cache $n$;
            \State Forward sample $s_n$ through input layers (last token embedding only);
            \State Place result in output message queue;
        \EndIf
   \EndFor
   \State Send stopping message to nodes;
   \State Decode samples with tokenizer;
   \State \textbf{return} List of generated pieces of text.
\end{algorithmic}
\end{algorithm}

We also report the secondary node operation algorithm in \ref{alg:secondary}.
Notice that the received messages also contain information about the sample ID $n$, allowing the node to select the correct KV cache when performing each forward pass.

\begin{algorithm}[tb]
   \caption{Secondary node main processing loop}
   \label{alg:secondary}
\begin{algorithmic}
    \State \textbf{input:} $n\_samples$
    \State Initialize sockets and message queues;
    \State Load model chunk and initialize model state;
    \State Initialize empty list of $n\_samples$ caches;
    \While{Stopping message not received}
        \State Extract sample $s_n$ from input queue;
        \If{$n$-th KV cache is None}
            \State \# \textit{First iteration for current sample}
            \State \verb|init_KV_cache(|$sample=s_n$\verb|)|;
        \EndIf
        \State Activate KV cache $n$;
        \State Forward sample $s_n$ through local model layers;
        \State Place result in output message queue.
   \EndWhile
\end{algorithmic}
\end{algorithm}

\section{\label{sec:eval}Performance Analysis}
\subsection{Setup}
This section contains the performance analysis of the MDI-LLM framework. First, we describe the LLM models that we used to validate our results. Our work is built on the LitGPT project \cite{litgpt-2023}, allowing us to choose among a large pool of LLM families.
In particular, we focus on deploying models with the features of the Llama 2 and Llama 3 families \cite{touvron2023llama}, i.e., decoder-only transformers using Rotary Positional Embeddings (RoPE), Grouped-Query Attention (GQA), and KV Caching.
To provide an equal comparison, we also performed tests using a small toy model we named ``NanoLlama'' (304 million parameters) with the same characteristics.
This model can fit on a single testbed device, providing a performance baseline.

Our testbed comprised three Nvidia Jetson TX2 boards (8 GB of shared RAM) \cite{jetsonSpecs} connected through a gigabit ethernet switch. These boards are a good example of typical edge devices, as their performance of $1.33$ TFLOPS is comparable to typical smartphones today. Thanks to pipeline parallelism, the final framework can achieve efficient generation and be scaled with minimal effort by adding devices to the network.
Another important aspect to keep in mind is that these boards do not support 16-bit operations.

To provide a fair comparison, we performed the tests on a toy LLM (``NanoLlama''), using the same architecture of Llama 2, but with a much lower number of parameters (304 million), which can fit completely on a single Nvidia Jetson from our testbed, trained on the ``Tiny Shakespeare'' data set \cite{tinyShakespeare}, which allowed it to run on a single testbed board. 
We consider the following model characteristics. Maximum context length is $2048$ tokens. Embedding dimension (activation length) is $1024$. The number of transformer blocks is $12$/ Model partition properties are; (i) Two nodes: n. starter blocks = 5, n. secondary blocks = 7. (ii) Three nodes: n. starter blocks = 2, n. secondary blocks = 5. The number of attention heads per Multi-Head Attention (MHA) layer is $16$. Vocabulary size is $32000$ tokens.


The other tests were performed on the TinyLlama Chat v1.0 model \cite{zhang2024tinyllama}, a 1.1 billion parameter model with the following features.
Maximum context length is  $2048$ tokens. Embedding dimension (activation length) is $2048$. Number of transformer blocks is $22$. Model partition: (i) Two nodes: n. starter blocks = 10, n. secondary blocks = 12. (ii) Three nodes: n. starter blocks = 6, n. secondary blocks = 8. Number of attention heads per Multi-Head Attention (MHA) layer is $32$. Vocabulary size is $32000$ tokens. 
%
%
This model cannot fit in the memory of a single board, as the overall VRAM/RAM requirements that take into account the model parameters (4.6 GB alone), all caches, the internal state initialized at runtime, and the Python libraries cannot be accommodated in 8 GB of shared memory (that is also used to accommodate the operating system).

\subsection{Token Generation Rates}

\begin{figure}[t!]
\begin{center}
\centerline{\includegraphics[width=\columnwidth]{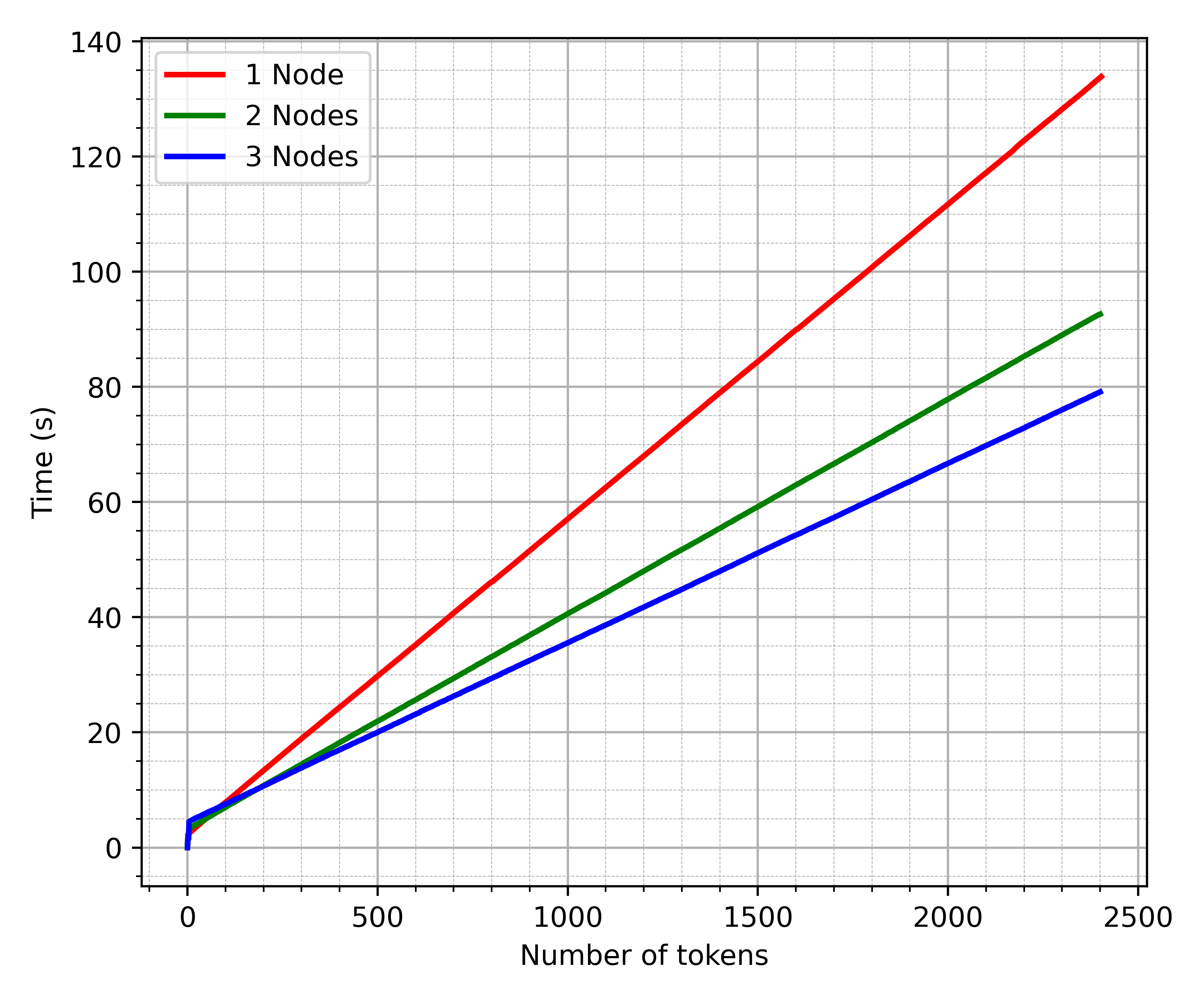}}
\caption{Time vs. generated tokens -- comparison on 300M parameters model for 3 generated samples, 800 tokens each.}
\label{perf-300m}
\end{center}
\vspace{-20pt}
\end{figure}

In Fig. \ref{perf-300m}, we compare the generation rates of the NanoLlama model producing three samples (800 tokens each) when deployed on one, two, and three devices by plotting the generation time versus the number of generated tokens.
The generation was initiated by providing a single ``\verb|\n|'' character to the model, i.e., a prompt containing one single character.
It is evident how increasing the number of nodes in the system makes it possible to achieve higher generation rates for a fixed number of desired tokens.

\begin{figure}[ht]
\begin{center}
\centerline{\includegraphics[width=\columnwidth]{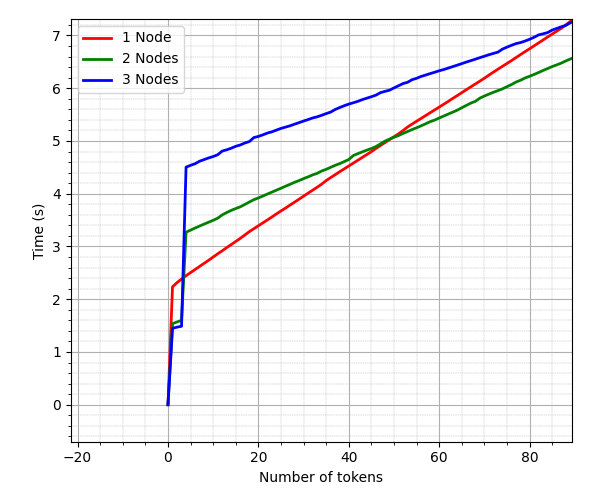}}
\caption{Behavior at the origin -- it is possible to notice the initial slowdown due to KV cache and internal state initialization.}
\label{detail-300m}
\end{center}
\vspace{-20pt}
\end{figure}

Fig. \ref{detail-300m} reports a zoom-in of the same plot at the beginning of the generation.
It is possible to observe an initial transient in which the generation temporarily proceeds at a lower rate.
This initial slowdown is due to the time required to initialize the local models' KV cache and internal state (handled by the PyTorch library), and it is possible to see that it increases proportionally with the number of nodes.
After all the nodes have processed each sample once, the generation rate reaches a stable value. The plot also provides relevant details, such as the effect of network jitter on the generation rate.

\subsection{Memory Usage}

Another fundamental aspect of applying MDI is reducing the memory usage on each device, allowing the LLM to be distributed over multiple nodes.
Table \ref{tab-mem-300m} compares the total (VRAM + RAM, the system memory is shared in the considered testbed devices) memory usage for NanoLlama 304M in the three considered settings.

\begin{table}[ht]
\caption{Memory usage in GB -- NanoLlama 304M}
\label{tab-mem-300m}
\vskip 0.15in
\begin{center}
\begin{small}
\begin{sc}
\begin{tabular}{cccc}
\toprule
N. nodes & Node 1 & Node 2 & Node 3 \\
\midrule
\textbf{1} & 2.15 & \textit{N/A} & \textit{N/A}\\
\textbf{2} & 1.76 & 1.76 & \textit{N/A} \\
\textbf{3} & 1.34 & 1.46 & 1.46 \\
\bottomrule
\end{tabular}
\end{sc}
\end{small}
\end{center}
\vskip -0.1in
\end{table}

We observe that the memory requirements can be significantly reduced with an increasing number of devices, with a law of diminishing returns similarly applying as the number of devices increases. For example, $1.76$ GB/device is needed for two nodes, while only $1.56$ GB/device is needed for three nodes.
Due to overheads, the total memory required increases as the number of devices increases. For example, the Python libraries alone require $450$ MB of memory, and, in the multi-node case, approximately $150$ to $200$ MB are used to run the HTTP server for coordination and for maintaining the communication channels used to transmit the model activations.
Another important point to consider is how KV caching and other stored elements, like the attention layer masks and the pre-computed RoPE values, increase the graphics memory usage at runtime.
Indeed, the model parameters as stored on disk only occupy $1.1$ GB of space. We believe that most of these overheads will likely disappear with an optimized implementation. Still, our MDI-LLM scheme provides significant memory usage reduction. 

\begin{table}[ht]
\caption{Memory usage in GB -- TinyLlama 1.1B}
\label{tab-mem-tinyllama}
\vskip 0.15in
\begin{center}
\begin{small}
\begin{sc}
\begin{tabular}{cccc}
\toprule
N. nodes & Node 1 & Node 2 & Node 3 \\
\midrule
\textbf{2} & 4.57 & 4.57 & \textit{N/A} \\
\textbf{3} & 3.26 & 3.26 & 3.26 \\
\bottomrule
\end{tabular}
\end{sc}
\end{small}
\end{center}
\vskip -0.1in
\end{table}

In Table \ref{tab-mem-tinyllama}, we report the memory usage for the two considered scenarios for the TinyLlama model.
A key distinction as compared with the previous model is that, here, the model is unable to fit to a single device.
Our MDI-LLM framework allows inference over the model by distributing it model over multiple devices that otherwise would not be able to run it.
Also, going from two to three devices reduces the per-device memory usage by $1.3$GB. The overhead is still present, as the total amount of memory used by the whole network increases slightly when adding nodes ($9.14$ GB for two nodes, $9.78$ GB for three). This increase is consistent with the size of the Python libraries and caches reported previously.

\section{\label{sec:conc}Conclusions}

In this paper, we presented MDI-LLM, a novel framework that enables the deployment of state-of-the-art large language models (LLMs) across low-power edge devices. To maximize inference efficiency, we introduced recurrent pipeline parallelism, a technique that minimizes idle time across devices and supports parallel generation of multiple text sequences. MDI-LLM also incorporates techniques such as KV caching and Grouped Query Attention (GQA), which is crucial for the efficiency of MDI-LLM. Our results demonstrate that MDI-LLM can harness the combined computational resources of multiple edge nodes to run models that exceed the memory constraints of individual devices—enabling LLM inference on cost-effective hardware. This work provides a foundation for scalable, model-distributed inference of transformer-based architectures in edge environments and paves the way for future research in multi-device, resource-constrained LLM deployment.

\bibliographystyle{IEEEtran}

\bibliography{IEEEabrv,references}

%
%

\end{document}